\documentclass[a4paper]{jpconf}
\usepackage{multirow}
\usepackage{array,booktabs}
\usepackage{diagbox}
\usepackage{balance}
\usepackage{xspace}
\usepackage{algorithm}
\usepackage{algpseudocode}
\usepackage{graphicx}
\usepackage{amsmath}
\usepackage{amssymb}  
\usepackage{subfigure}
\usepackage{adjustbox}
\usepackage{romannum}
 \usepackage[final]{hyperref}
 \hypersetup{
     colorlinks=true,
     linkcolor=blue,
     filecolor=magenta,
     urlcolor=blue,
     citecolor=black
 }

\usepackage{subcaption}
\usepackage{caption}
\begin{document}
\title{Enhancing 3D Object Detection by Using Neural Network with Self-adaptive Thresholding}

\author{Houze Liu$^{1,\dagger
}$, Cangqing Wang$^{1,\star}$,Xiaoan Zhan$^2$, Haotian Zheng$^3$, Chang Che$^*$}

\address{$^{1,\dagger
}$New York University, NY, USA}

\address{$^{1,\star}$Independant researcher}
\address{$^{2}$New York University, NY, USA}
\address{$^{3}$New York University, NY, USA}
\address{$^{*}$George Washington University, DC, USA}
\ead{$^{1,\star}$hl2979@nyu.edu, $^{1,\star}$cangqingwang@gmail.com, $^{2}$zgaq2502@gmail.com, $^{3}$hz2687@nyu.edu, $^{*}$cche57@gwmail.gwu.edu}

\begin{abstract}
Robust 3D object detection remains a pivotal concern in the domain of autonomous field robotics. Despite notable enhancements in detection accuracy across standard datasets, real-world urban environments, characterized by their unstructured and dynamic nature, frequently precipitate an elevated incidence of false positives, thereby undermining the reliability of existing detection paradigms. In this context, our study introduces an advanced post-processing algorithm that modulates detection thresholds dynamically relative to the distance from the ego object. Traditional perception systems typically utilize a uniform threshold, which often leads to decreased efficacy in detecting distant objects. In contrast, our proposed methodology employs a Neural Network with a self-adaptive thresholding mechanism that significantly attenuates false negatives while concurrently diminishing false positives, particularly in complex urban settings. Empirical results substantiate that our algorithm not only augments the performance of 3D object detection models in diverse urban and adverse weather scenarios but also establishes a new benchmark for adaptive thresholding techniques in field robotics.
\end{abstract}
\begin{keywords}
Neural Network, 3D Detection, Computer Vision,  Thresholding 
\end{keywords}

\section{Introduction}
Robust 3D object detection within dynamic and unstructured environments is crucial for the effective operation of field robotics, particularly in urban settings characterized by their inherent complexity and variability. Urban landscapes, while seemingly structured, are fraught with dynamic elements such as moving objects and pedestrians, alongside unpredictable static obstacles like foliage and street signage. These factors, compounded by sensor noise and the additional challenges posed by adverse weather conditions such as fog and rain, significantly exacerbate the risk of false positives. Such inaccuracies can lead to severe operational disruptions~\cite{che2023enhancing,zhou2023distributed,zou2023joint}, including abrupt vehicular stops and traffic collisions~\cite{li2024ddn}. Consequently, the development of an advanced perception module that effectively minimizes false positives in urban environments represents a critical research frontier in the domain of field robotics~\cite{che2024intelligent,gao2023autonomous}.

LiDAR technology, with its high-resolution capabilities, remains foundational in 3D object detection, facilitating the generation of precise 3D environmental data crucial for robust perception~\cite{zhou2024machine,pan2023navigating,chen2024comprehensive}. Employing 3D point clouds, LiDAR-based detection models enable the prediction of an object’s class~\cite{yu2024similarity}, location, and confidence scores. These initial predictions are subsequently refined through confidence score-based post-processing, typically utilizing a singular threshold hyper-parameter~\cite{zhang2020manipulator}. However, detection accuracy is contingent upon the distance of the object from the sensor, influenced by sensor resolution, range, and the diversity of the training datasets. Objects proximal to the ego-object exhibit heightened accuracy and confidence owing to denser point clouds, whereas those at a distance manifest reduced recall and confidence due to sparser point clouds. This disparity underscores the necessity for autonomous systems navigating real roads to adapt their emphasis—prioritizing precision for nearer objects and recall for those farther away~\cite{yu2024stochastic, zhibin2019labeled}. Relying on a singular post-processing threshold proves inadequate for managing the varied demands of diverse real-world environments.

\begin{figure*}[t!]
    \centering
    \includegraphics[width=1\textwidth]
    {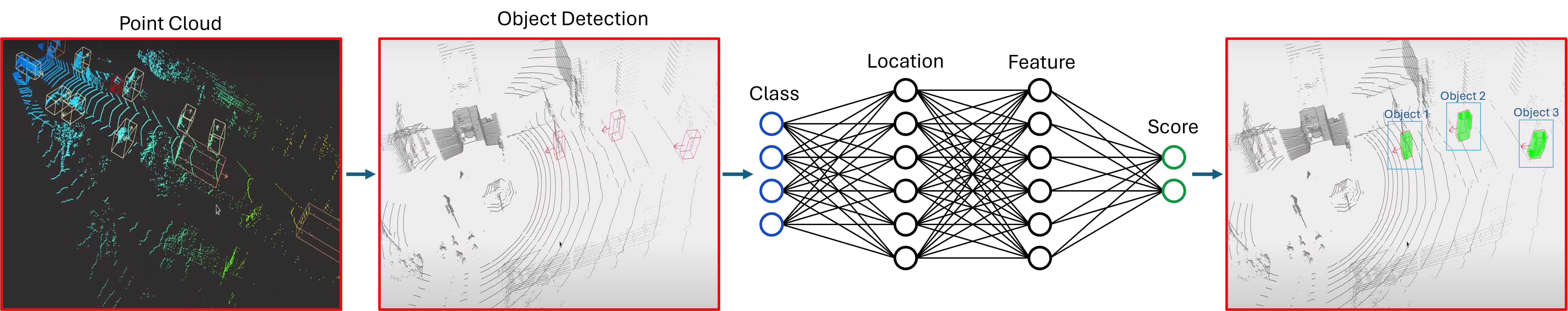}
    \caption{
    The Nerual-Network framework of the proposed algorithm. Traditional 3D detection models typically employ a static, binary-threshold approach for post-processing. In contrast, our method utilizes Neural Network thresholding, which dynamically adjusts thresholds based on specific environmental inputs and detection contexts, thereby enhancing detection accuracy and reliability.
    }
    \label{fig:1}
    \vspace{-0.3cm}
\end{figure*}

While prior research on post-processing algorithms has demonstrated advancements in adaptive thresholding techniques primarily within 2D image processing domains~\cite{zou2022unified}, these methodologies have not been wholly adapted to the nuanced requirements of 3D object detection~\cite{wang2024intelligent}. Addressing this gap, our study introduces a sophisticated adaptive thresholding algorithm tailored explicitly for 3D object detection in autonomous mobile systems. This innovation integrates seamlessly into existing frameworks without necessitating additional training or complicating the detection architecture~\cite{wang2024adapting}. By finely tuning the balance between reducing false positives for proximate objects and diminishing missed detections for distant objects, our approach significantly bolsters the performance of 3D object detection across both controlled datasets~\cite{li2024enhancing} and complex real-world environments~\cite{zou2023multidimensional}. This enhancement crucially contributes to safer and more effective autonomous navigation, as evidenced by our qualitative analysis in Fig.\ref{fig:1}.

\section{Related work}
\subsection{Thresholding in Object Detection}
Previous studies on adaptive thresholding, specifically targeting moving object detection in 2D images, have primarily focused on enhancing detection accuracy through distance-based methodologies. achieved this by identifying objects within images and estimating their states and distances through pixel comparison across successive frames. Conversely, applied adaptive thresholding to set thresholds essential for differentiating between moving and stationary objects during depth estimation is crucial for translating 2D images into 3D representations. This technique dynamically adjusts thresholds based on the proximity of paired image features, thereby aiding in the determination of an object’s motion status. However, these approaches are inherently limited to 2D image contexts and do not extend to the complex requirements of 3D object detection in autonomous driving. Autonomous systems necessitate a robust 3D spatial representation to accurately recognize and react to various objects and conditions.
\subsection{LiDAR based 3D Object Detection}
3D object detection encompasses a range of modalities, including LiDAR-based, camera-based, radar-based, and sensor-fusion-based techniques. This paper focuses exclusively on LiDAR-based 3D object detection strategies, segmenting them into three primary categories: voxel-based, pillar-based, and raw point cloud-based approaches.

Within these classifications, VoxelNet employs a technique of aggregating and downsampling 3D point clouds into voxels to construct a robust feature representation, subsequently executing object detection through 3D convolution operations supported by a Region Proposal Network (RPN). SECOND enhances voxel processing by introducing a computation strategy tailored for the sparse nature of 3D environmental data. PointPillars transitions to a pillar-based approach, utilizing 2D convolution to optimize computational and memory efficiency while maintaining robust detection capabilities. PointRCNN leverages raw point clouds to differentiate objects from the background, utilizing this distinct classification for precise object detection. PV-RCNN combines the advantages of both raw point clouds and voxels to preserve the intrinsic details of point clouds, despite the computational demands\cite{che2024integrating}, achieving significant efficiency and detection accuracy. These methods predominantly employ anchor-based detection frameworks. In contrast, CenterNet introduces an anchor-free detection paradigm, challenging the conventional anchor-based approaches by promoting a less constrained detection framework.

Although these models demonstrate efficacy in controlled, typical road environments, they often falter in handling the more complex and variable conditions encountered on real urban roads. On these roads, varying weather conditions such as rain, snow, and fog can introduce significant noise into sensor data, complicating the detection process. Additionally, urban elements like bushes or road signs may be erroneously identified as objects or other obstacles, leading to an increased rate of false positives. To address these challenges, we propose an advanced method that enables autonomous objects to adaptively manage false positives, thereby enhancing stability and reliability in real-world urban navigation.

\section{Self-Adaptation Thresholding with Neural-Net work}
The analytical results, utilizing differentiated single thresholds—0.5 for proximal distances (below 40 meters) and 0.3 for extended distances (above 40 meters)—enabled precise computation of the standard deviation at 10-meter intervals, as illustrated in Fig.. Upon scrutinizing the trends in confidence scores depicted in  Fig., it became evident that the variability of confidence scores with respect to distance adheres most closely to a quadratic functional form. This observation facilitated the derivation of a distance-based adaptive thresholding formula, as elaborated in EQ.. This formula not only accommodates the fluctuations in detection confidence across varying distances but also optimizes the threshold settings for enhanced detection accuracy in diverse operational contexts.

Consider a point cloud represented by $\boldsymbol{W} = [\mathbf{w}_1, \dots, \mathbf{w}_k]^\mathsf{T} \in \mathbb{R}^{{k}\times 3}$ and its associated chromatic data given by $\mathbf{Z} = [\mathbf{z}_1, \dots, \mathbf{z}_k]^\mathsf{T} \in \mathbb{R}^{{k}\times 3}$. Here, each vector $\mathbf{w}_k$ encodes the spatial coordinates in the object-centric frame, whereas $\mathbf{z}_k$ corresponds to the RGB values of each point. Leveraging these datasets, the training of a detection algorithm can be conceptualized as a supervised colorization task, wherein each pair $(\mathbf{W}, \mathbf{Z})$ serves as a distinct training sample.

\begin{equation}
\min_{\boldsymbol{\zeta}, \boldsymbol{\xi}} \sum_{\boldsymbol{W}, \boldsymbol{Z} \in \mathcal{S}_p}\left\|\mathcal{U}_+ \otimes \mathcal{F}_{\boldsymbol{\zeta}}(\boldsymbol{W})-\boldsymbol{Z}\right\|=\min _{\boldsymbol{k_1}, \boldsymbol{k_2}} \sum_{\boldsymbol{W}, \boldsymbol{Z} \in \boldsymbol{S}_p} \sum_k\left\|\widetilde{\boldsymbol{z}}_k-\boldsymbol{z}_k\right\|,
\label{eq1}
\end{equation}
where $\mathcal{S}_p$ denote the training data and $\hat{\boldsymbol{z}}_k$ denote the estimated data of $\boldsymbol{Z}$ set.


The analytical outcomes, employing single thresholds of 0.5 for proximal distances and 0.3 for more extended distances , facilitated the computation of the standard deviation across intervals of 10m, as described in Fig.\ref{fig:1}. After an examination of confidence score trends depicted in Fig.\ref{fig:1}, it was inferred that a quadratic function most appropriately captures the variability of confidence scores across distances~\cite{zhou2023semantic}, leading to the formulation of the distance-based adaptive thresholding equation as

\begin{equation}
Q\left(\sigma_{x y}, m_{x y}\right)= \begin{cases}\text { TRUE } & \text { if } f(x, y)>\alpha \sigma_{x y} \text { AND } f(x, y)>\beta m_{x y} \\ \text { FALSE } & \text { otherwise }\end{cases}
\label{eq2}
\end{equation}
where $m_{{x y}}=\sum_{i=0}^{k} r_i p_{{x y}}\left(r_i\right)$ and $\sigma_{x y}=\sum_{i=0}^{k}\left(r_i-m_{{x y}}\right) p_{{x y}}\left(r_i\right)$. The term $p_{xy}$ denotes the normalized histogram value corresponding to the intensity $r_i$, which quantifies the distance from the ego-object to the observed object in meters. The coefficients $\alpha$ and $\beta$ shape the quadratic function that encapsulates the confidence score's variation as a function of distance. The parameter $m_{xy}$ establishes the maximum operational distance for the algorithm. The variable $\sigma_{xy}$, indicative of the LiDAR's detection range and the efficacy of the 3D object detection models, may fluctuate. To mitigate an excessive reduction in the confidence threshold beyond $\sigma_{xy}$—thereby curbing the surge in false positives—a constant $k$ is utilized. These parameters facilitate the adaptation of the algorithm to diverse 3D object detection frameworks, including but not restricted to the PointPillars model.

\begin{equation}
\mathcal{Q}_\Delta=\frac{\sum_{i=d}^{d+1} \mathcal{Q}_i}{N_d}
\label{eq3}
\end{equation}
In each respective interval, the aggregate confidence scores of discerned entities are normalized by the count of detections to ascertain the mean confidence metric. Concurrently, the standard deviation of these scores, as delineated in Eq.\ref{eq3}, is computed relative to the interval's mean confidence. Subsequently, a quadratic regression is formulated by refining the sextet of mean confidence values procured from Eq.\ref{eq3} in accordance with the standard deviations stipulated in Eq.\ref{eq2}.
\section{Experiments Results}
Two disparate datasets $\mathcal{D}_1,\mathcal{D}_2$ were employed to rigorously evaluate our proposed algorithms: $\mathcal{D}_1$ is for the sidewalk at night, and $\mathcal{D}_2$ is from the woods on the edge of the city. We use the public data set collected from OS2 LiDAR sensors. To augment the dataset's variability further, a drone outfitted with OS2 LiDAR sensors was deployed. This initiative was aimed at assembling a comprehensive corpus of data that encapsulates a diverse array of meteorological conditions.

\begin{table}[!htb]
\caption{\small Comparison with different methods using our training data set.}
\centering

\setlength{\tabcolsep}{4px}
\begin{tabular}{lccccc}
\toprule
Algorithem & Arch & Dataset & 15\% & 25\% & 35\% \\ \midrule
Otsu & UNet & $\mathcal{D}_1$ & 73.7 & 76.4 & 80.1 \\
Nick & UNet & $\mathcal{D}_1$ & 62.8 & 65.9 & 71.1\\
Bernsen & UNet & $\mathcal{D}_1$ & 67.3 & 72.5 & 74.9 \\
Phansalkar & UNet & $\mathcal{D}_1$ & 68.4 & 71.6 & 74.8 \\
Bradley & UNet & $\mathcal{D}_1$ & 65.7 & 69.6 & 70.8 \\
\textbf{Ours} & UNet + NN & $\mathcal{D}_1$ + $\mathcal{D}_2$ & 60.7 & 62.4 & 69.1 \\ 
\textbf{Ours} & UNet + NN & $\mathcal{D}_1$ + $\mathcal{D}_2$ & 61.1 & 67.0  & 72.4 \\ \hline
\bottomrule
\end{tabular}
\label{tab1}
\end{table}

\begin{figure}[!htb]
    \centering
    \includegraphics[width=1\columnwidth]{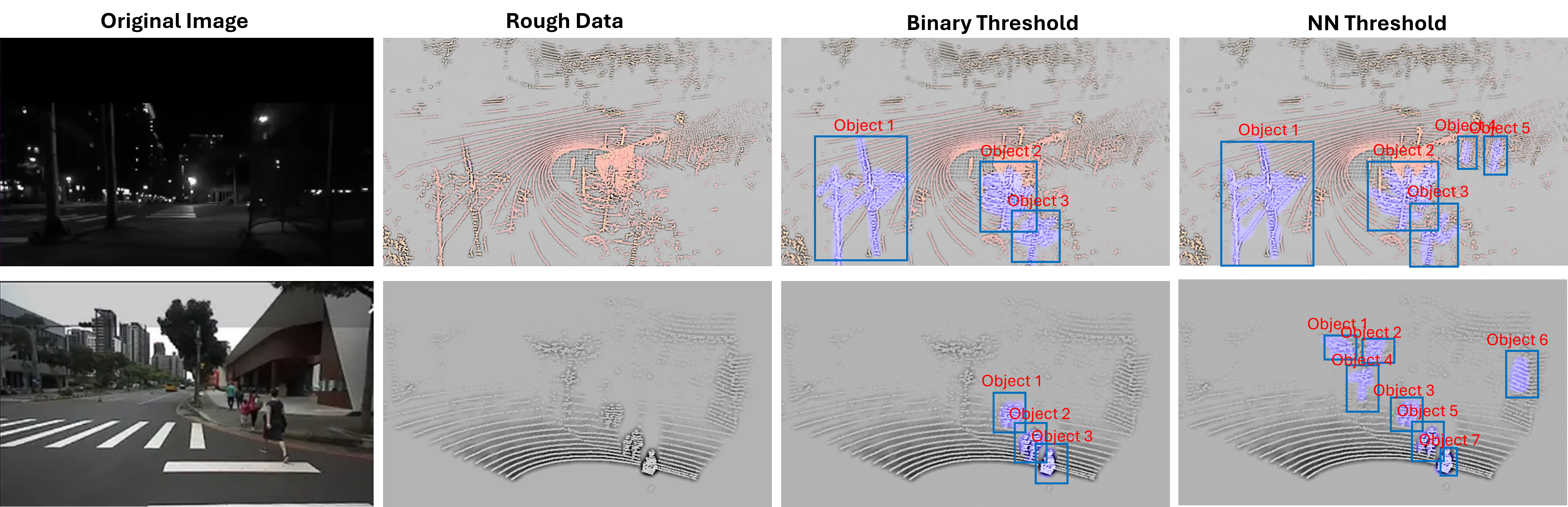}
    \caption{\small Results for comparison of our method and traditional method. Our algorithm significantly enhances the efficacy of 3D object detection models in complex urban environments characterized by adverse weather conditions such as fog and rain. By effectively minimizing false positives and accurately discerning objects from misleading point clouds generated by inclement weather, the algorithm improves the precision of object detection. This advancement fosters safer navigation for autonomous objects by ensuring more reliable and accurate perception under challenging conditions.} 
\label{fig2}
\end{figure}

Figure \ref{fig2} illustrates different datasets collected during significant fog in Baltimore. The RGB imagery in Figure \ref{fig2} depicts conditions characterized by heavy precipitation~\cite{wang2024jointly}. Despite the presence of objects in the left lane, it is noted that neither the driving lane directly ahead nor the right lane contains any objects. Different algorithm comparison results are shown in Table.\ref{tab1}, where vehicular false positives are also attributed to bushes on the right side of the roadway. Our algorithm has been refined to enhance precision at proximal distances specifically to address and reduce false positives caused by rain and vegetative obstructions, while still maintaining accurate detection of obstacle situated in different environment.

\section{Conclusion}
This investigation introduced an advanced adaptive thresholding algorithm designed for post-processing applications within three-dimensional object detection models, which are crucial to the perception modules of autonomous objects—a central element of Field Robotics. Through methodical comparative analyses conducted using an accessible dataset, we have established that our algorithm substantially enhances the robustness of the perception module. It adeptly reduces false positives by dynamically modulating thresholds based on the object's distance from detected objects. Additionally, there is potential for this algorithm to evolve into a learning-based model that can adapt in real time to a multitude of driving scenarios, thereby enhancing its applicability and performance in the field of autonomous driving.

\section*{References}
\bibliography{ref}

\begin{thebibliography}{10}

\bibitem{che2023enhancing}
Chang Che, Qunwei Lin, Xinyu Zhao, Jiaxin Huang, and Liqiang Yu.
\newblock Enhancing multimodal understanding with clip-based image-to-text transformation.
\newblock In {\em Proceedings of the 2023 6th International Conference on Big Data Technologies}, pages 414--418, 2023.

\bibitem{zhou2023distributed}
Lisang Zhou, Meng Wang, and Ning Zhou.
\newblock Distributed federated learning-based deep learning model for privacy mri brain tumor detection.
\newblock {\em Journal of Information, Technology and Policy}, pages 1--12, 2023.

\bibitem{zou2023joint}
Zhibin Zou, Maqsood Careem, Aveek Dutta, and Ngwe Thawdar.
\newblock Joint spatio-temporal precoding for practical non-stationary wireless channels.
\newblock {\em IEEE Transactions on Communications}, 71(4):2396--2409, 2023.

\bibitem{li2024ddn}
Mingrui Li, Jiaming He, Guangan Jiang, and Hongyu Wang.
\newblock Ddn-slam: Real-time dense dynamic neural implicit slam with joint semantic encoding.
\newblock {\em arXiv preprint arXiv:2401.01545}, 2024.

\bibitem{che2024intelligent}
Chang Che, Haotian Zheng, Zengyi Huang, Wei Jiang, and Bo~Liu.
\newblock Intelligent robotic control system based on computer vision technology.
\newblock {\em arXiv preprint arXiv:2404.01116}, 2024.

\bibitem{gao2023autonomous}
Longsen Gao, Giovanni Cordova, Claus Danielson, and Rafael Fierro.
\newblock Autonomous multi-robot servicing for spacecraft operation extension.
\newblock In {\em 2023 IEEE/RSJ International Conference on Intelligent Robots and Systems (IROS)}, pages 10729--10735. IEEE, 2023.

\bibitem{zhou2024machine}
Lisang Zhou, Ziqian Luo, and Xueting Pan.
\newblock Machine learning-based system reliability analysis with gaussian process regression.
\newblock {\em arXiv preprint arXiv:2403.11125}, 2024.

\bibitem{pan2023navigating}
Xueting Pan, Ziqian Luo, and Lisang Zhou.
\newblock Navigating the landscape of distributed file systems: Architectures, implementations, and considerations.
\newblock {\em Innovations in Applied Engineering and Technology}, pages 1--12, 2023.

\bibitem{chen2024comprehensive}
Feiyang Chen, Ziqian Luo, Lisang Zhou, Xueting Pan, and Ying Jiang.
\newblock Comprehensive survey of model compression and speed up for vision transformers.
\newblock {\em Journal of Information, Technology and Policy}, pages 1--12, 2024.

\bibitem{yu2024similarity}
Liqiang Yu, Bo~Liu, Qunwei Lin, Xinyu Zhao, and Chang Che.
\newblock Similarity matching for patent documents using ensemble bert-related model and novel text processing method.
\newblock {\em Journal of Advances in Information Technology}, 15(3), 2024.

\bibitem{zhang2020manipulator}
Yufeng Zhang, Xue Wang, Longsen Gao, and Zongbao Liu.
\newblock Manipulator control system based on machine vision.
\newblock In {\em International Conference on Applications and Techniques in Cyber Intelligence ATCI 2019: Applications and Techniques in Cyber Intelligence 7}, pages 906--916. Springer, 2020.

\bibitem{yu2024stochastic}
Liqiang Yu, Chen Li, Bo~Liu, and Chang Che.
\newblock Stochastic analysis of touch-tone frequency recognition in two-way radio systems for dialed telephone number identification.
\newblock {\em arXiv preprint arXiv:2403.15418}, 2024.

\bibitem{zhibin2019labeled}
ZOU Zhibin, SONG Liping, and Cheng Xuan.
\newblock Labeled box-particle cphd filter for multiple extended targets tracking.
\newblock {\em Journal of Systems Engineering and Electronics}, 30(1):57--67, 2019.

\bibitem{zou2022unified}
Zhibin Zou, Maqsood Careem, Aveek Dutta, and Ngwe Thawdar.
\newblock Unified characterization and precoding for non-stationary channels.
\newblock In {\em ICC 2022-IEEE International Conference on Communications}, pages 5140--5146. IEEE, 2022.

\bibitem{wang2024intelligent}
Cangqing Wang.
\newblock Intelligent agricultural greenhouse control system based on internet of things and machine learning.
\newblock {\em arXiv preprint arXiv:2402.09488}, 2024.

\bibitem{wang2024adapting}
Cangqing Wang, Yutian Yang, Ruisi Li, Dan Sun, Ruicong Cai, Yuzhu Zhang, Chengqian Fu, and Lillian Floyd.
\newblock Adapting llms for efficient context processing through soft prompt compression.
\newblock {\em arXiv preprint arXiv:2404.04997}, 2024.

\bibitem{li2024enhancing}
Chen Li, Haotian Zheng, Yiping Sun, Cangqing Wang, Liqiang Yu, Che Chang, Xinyu Tian, and Bo~Liu.
\newblock Enhancing multi-hop knowledge graph reasoning through reward shaping techniques.
\newblock {\em arXiv preprint arXiv:2403.05801}, 2024.

\bibitem{zou2023multidimensional}
Zhibin Zou and Aveek Dutta.
\newblock Multidimensional eigenwave multiplexing modulation for non-stationary channels.
\newblock In {\em GLOBECOM 2023-2023 IEEE Global Communications Conference}, pages 2524--2529. IEEE, 2023.

\bibitem{che2024integrating}
Chang Che, Zengyi Huang, Chen Li, Haotian Zheng, and Xinyu Tian.
\newblock Integrating generative ai into financial market prediction for improved decision making.
\newblock {\em arXiv preprint arXiv:2404.03523}, 2024.

\bibitem{zhou2023semantic}
Yiming Zhou, Ahmad Osman, Marc Willms, Albrecht Kunz, Selina Philipp, Janine Blatt, and Simon Eul.
\newblock Semantic wireframe detection.
\newblock 2023.

\bibitem{wang2024jointly}
Han Wang, Yiming Zhou, Eduardo Perez, and Florian Roemer.
\newblock Jointly learning selection matrices for transmitters, receivers and fourier coefficients in multichannel imaging.
\newblock {\em arXiv preprint arXiv:2402.19023}, 2024.

\end{thebibliography}
\end{document}